\documentclass{article}
\usepackage{ijcai21}

\usepackage{times}

\usepackage{soul}
\usepackage{url}
\usepackage[hidelinks]{hyperref}
\usepackage[utf8]{inputenc}
\usepackage[small]{caption}
\usepackage{graphicx}
\usepackage{amsmath}
\usepackage{booktabs}
\usepackage{tikz}
\newcommand*\circled[1]{\tikz[baseline=(char.base)]{
            \node[shape=circle,draw,inner sep=0.6pt] (char) {#1};}}
\usepackage{makecell}
\usepackage{multirow, array}
\usepackage{subfig}
\usepackage[switch]{lineno}

\usepackage{xspace}
\newcommand{\projectname}{CANAO\xspace}
\usepackage{color}
\usepackage{pifont}

\usepackage{amsmath}
\usepackage[noend, ruled]{algorithm2e}

\title{A Compression-Compilation Framework for On-mobile Real-time BERT Applications}

\author{
Wei Niu$^1$\thanks{These authors contributed equally}\and
Zhenglun Kong$^2$\footnotemark[1]\and
Geng Yuan$^2$\and
Weiwen Jiang$^3$\and
Jiexiong Guan$^1$\and\\
Caiwen Ding$^4$\and
Pu Zhao$^2$\and
Sijia Liu$^5$\and
Bin Ren$^1$\And
Yanzhi Wang$^2$
\affiliations
$^1$College of William and Mary\and
$^2$Northeastern University\and
$^3$University of Notre Dame\and\\
$^4$University of Connecticut\and
$^5$MIT-IBM Watson AI Lab, IBM Research\\
\emails
    \{wniu,jguan, bren\}@email.wm.edu, 
    \{kong.zhe,yuan.geng,zhao.pu, yanz.wang\}@northeastern.edu, \\
    wjiang2@nd.edu, 
    caiwen.ding@uconn.edu, 
    sijia.liu@ibm.com
}

\begin{document}

\maketitle

\begin{abstract}
Transformer-based deep learning models have increasingly demonstrated high accuracy on many natural language processing (NLP) tasks. In this paper, we propose a compression-compilation co-design framework that can guarantee the identified model to meet both resource and real-time specifications of mobile devices. Our framework applies a compiler-aware neural architecture optimization method (\projectname), which can generate the optimal compressed model that balances both accuracy and latency. We are able to achieve up to 7.8$\times$ speedup compared with TensorFlow-Lite with only minor accuracy loss.
We present two types of BERT applications on mobile devices: Question Answering (QA) and Text Generation. Both can be executed in real-time with latency as low as 45ms. Videos for demonstrating the framework can be found on
\url{https://www.youtube.com/watch?v=_WIRvK_2PZI}
\end{abstract}

\section{Introduction}

Pre-trained large-scale language models such as BERT~\cite{devlin2018bert}, XLNet~\cite{yang2019xlnet}, RoBERTa~\cite{liu2019roberta}, and GPT-2~\cite{Radford2019LanguageMA} have substantially advanced the state-of-the-art across a wide spectrum of NLP tasks. 
With the increasing popularity of mobile AI applications and the concerns of information security and privacy, it is desirable to deploy these well-trained models on edge devices, and furthermore, to meet real-time requirements. However, these models often consist of hundreds (or even thousands) of computation layers and hundreds of millions of parameters. 
Therefore, how to accommodate the large and extremely deep models, such as BERT to edge device becomes an imminent problem. 

\begin{figure}[t!]
    \centering
    \includegraphics[width=0.85 \columnwidth]{ 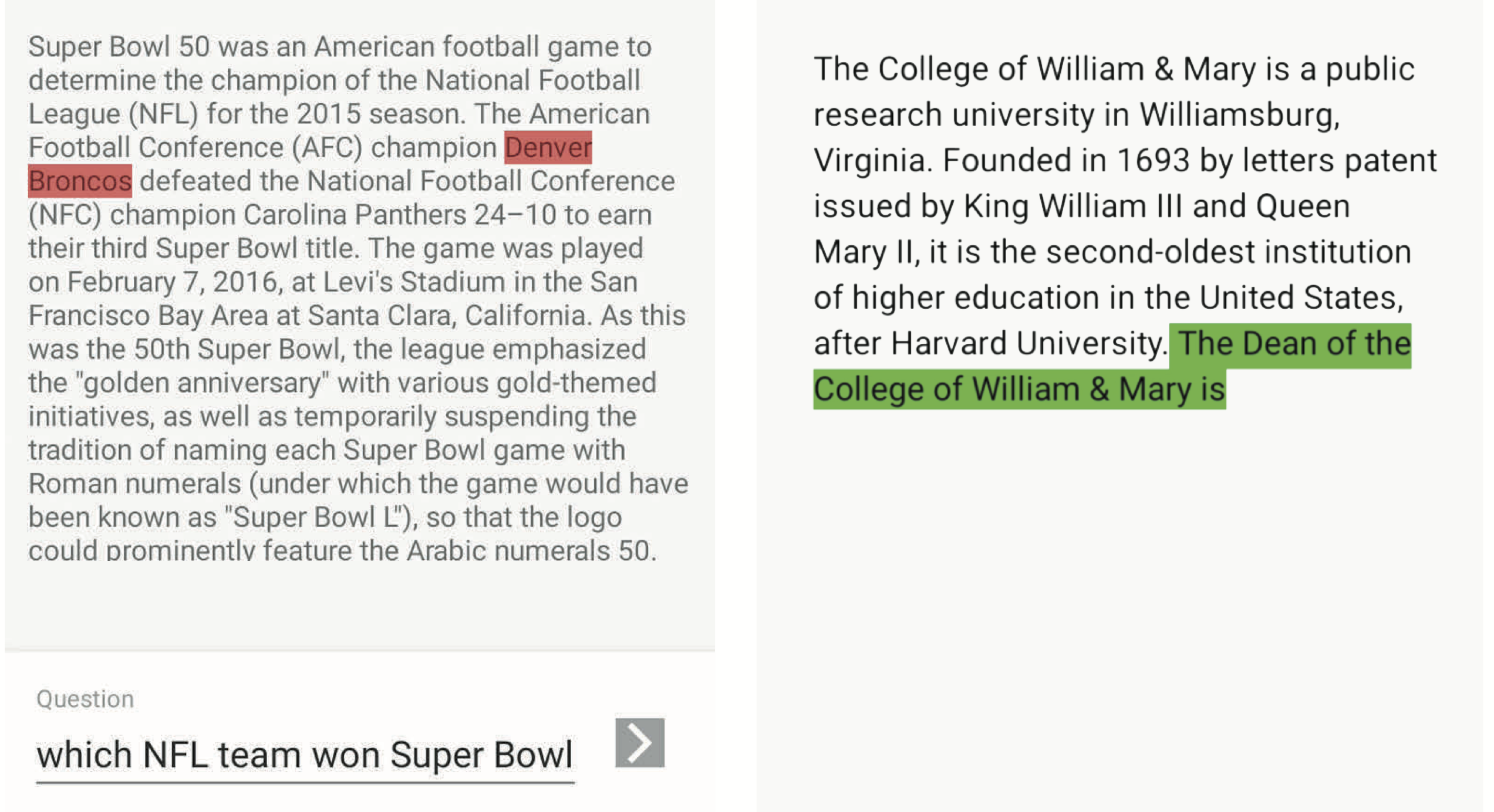}
    \vskip -0.5em
    \caption{
        Screenshot of the real-time BERT application on smartphone. Left: Question Answering task. Type a random question that is related to the paragraph, it will automatically highlight the answer in the test. Right: Text Generation task. Given a starting sentence, it can automatically generate new sentences by word.  }\label{fig:demo}
\end{figure}

There have been some efforts to compress the BERT model while maintaining the accuracy for downstream NLP tasks. MobileBERT~\cite{Sun_2020} is able to reduce the memory requirement, but there is still a considerable execution overhead due to a large number of computation units, thus leading to high inference latency.
Moreover, the large number of model layers also brings challenges in compiling models to mobile devices. To the best of our knowledge, only TensorFlow-Lite (TFLite)~\cite{TensorFlow-Lite} supports deploying BERT models on mobile CPU (not on mobile GPU), while no other frameworks can even support BERT models on mobile CPU. 


In this paper, we propose a compression-compilation co-design framework to optimize the structures of BERT variants for mobile devices. This is the first framework that involves compiler optimizations in the architecture search loop, aiming to co-optimize the model accuracy and computation resource usage. We also propose a highly effective layer fusion method to reduce intermediate results to achieve lower latency on both mobile CPU and GPU. Our framework outperforms the state-of-the-art framework, TFLite, by up to 7.8$\times$ speedup. Thus achieving the least latency while executing on mobile devices. We will release our model and framework.

\begin{figure*}[t]
    \centering
    \includegraphics[width=0.9 \textwidth]{ 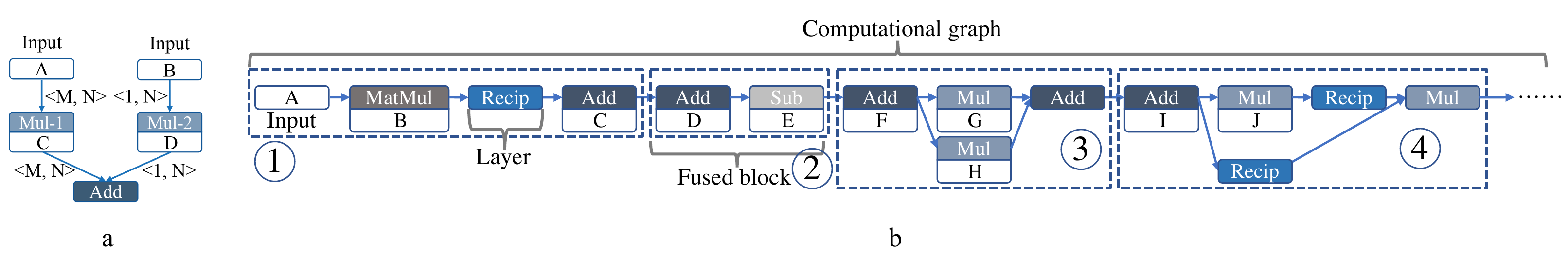}
    \vskip -0.5em
    \caption{a. Fusion example demonstrated in Figure~\ref{fig:codegen_example}. b. Sample fusion candidates for a computational graph section with an input (marked with A). Each layer has an input either from the previous layer/layers or from its weights, marked with other alphabets. Each number (from 1 to 4) denotes a fusion candidate (or fused block) based on mathematical properties.
    }\label{fig:fusion-pattern}
\end{figure*}
\section{Framework Design}

\begin{figure}[]
    \centering
    \setlength{\belowcaptionskip}{-5pt}
    \includegraphics[width=0.85 \columnwidth]{ 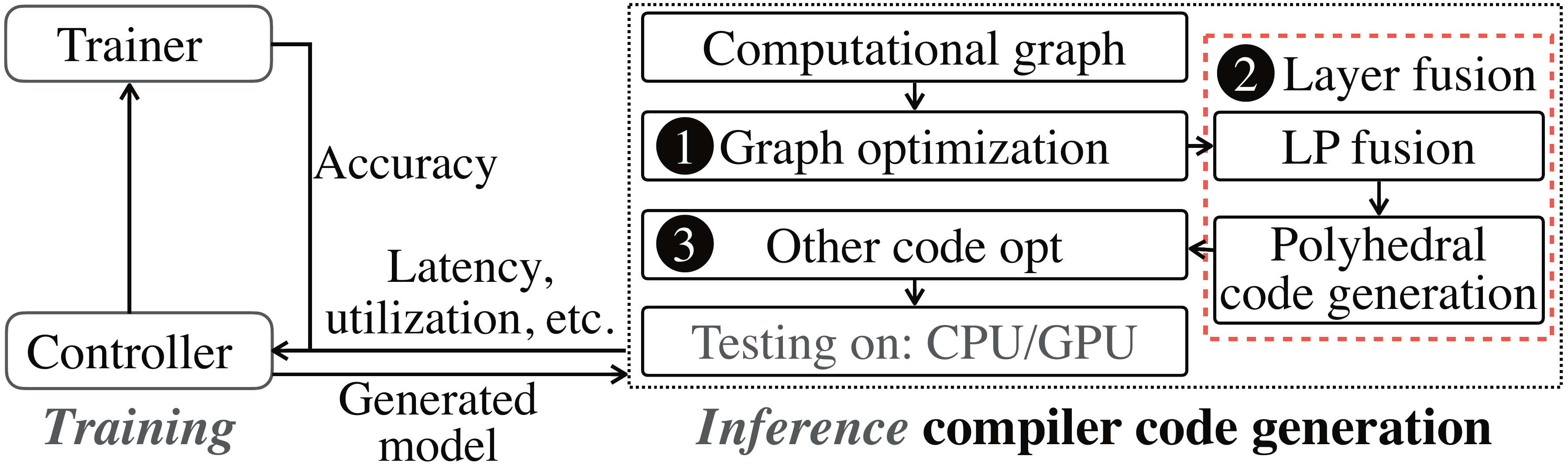}
    \vskip -0.5em
    \caption{
        Overview of compiler-aware neural architecture optimization framework.    }\label{fig:nas-compiler-overview}
\end{figure}

There are two processes in \projectname: {\it training} and {\it compiler code generation} (as shown in Figure~\ref{fig:nas-compiler-overview}). 
The training process includes a controller and a trainer.
The controller predicts/generates the model hyperparameters (i.e., network architecture); the trainer trains the predicted model and (quickly) evaluates its accuracy by fine-tuning the model to downstream tasks.
The compiler code generation process takes the predicted model and returns execution information (e.g. number of fused layers, latency, CPU/GPU utilization).
The execution information together with the model accuracy from the training process will be feedback to 
the controller to improve the prediction of neural architectures. After the compiler-aware NAS, the generated codes by our optimized compiler will be deployed for mobile CPU/GPU executions.

\textbf{For the training process}, the controller generates the architectural hyperparameters of neural networks. This includes two phases: 1) The determination of the number of transformer blocks; 2) The optimization of size for each layer. We find that layer number affects the accuracy the most for BERT related models, thus it should be the first thing we determine when searching the optimized model architecture. Then we optimize the layer size by considering both inference latency and model accuracy, which are set as reward signals to feedback to the controller. The controller serves to find the optimal architecture by maximizing the expected reward.

\textbf{The compiler code generation process} includes three steps: 1) Generate a computational graph from the controller-generated model and apply multiple optimizations on this graph.
2) Employ a novel compiler-based layer fusion optimization to further improve execution performance. This plays a key role in achieving better hardware efficiency.
3) Employ code generation and optimization to generate and further optimize the inference code. The generated inference code is tested on mobile devices. 
According to the feedback from the device side, the controller makes a better tradeoff between model accuracy and latency.

\subsection{Controller Architecture Search}\label{sec:NAS}
Our search space includes the number of layers, hidden layer size, and intermediate embedding size of the feedforward layers. We apply the recurrent neural network for searching the model architecture in the Controller. The recurrent network can be trained with a policy gradient method to maximize the expected reward of the sampled architectures. 
The accuracy and latency are used as the reward signal to feedback to the controller, which is trained by using the reinforcement learning method to explore the architecture. 
Our framework can search for a desirable model that achieves a good balance between accuracy and latency, preventing from searching the architecture manually.

\subsection{Compiler Code Generation}\label{sec:compiler-opt}



This section introduces our compiler optimizations that optimize the latency reward for the feedback.  
More specifically, it offers us multiple optimizing opportunities, e.g., reducing intermediate results, and eliminating unnecessary computations by analyzing the computation pattern. There are two phases for layer fusion: Lightweight Polynomial-based Layer Fusion (LP-Fusion) and Polyhedral-based Code Generation.
\subsubsection{LP-Fusion}
We identify all fusion candidates in a model based on two kinds of properties in the polynomial calculation: {\em computation laws} (i.e., associative, commutative, and distributive)  and {\em data access patterns}. 


Fig.~\ref{fig:fusion-pattern}b shows four fusion candidates (or fused blocks) for a computational graph. Layer fusion reduces not only the memory consumption of intermediate results, but also the number of operators. Take 
Fig.~\ref{fig:fusion-pattern}b\circled{3} for example, without layer fusion, the computation function is defined as:
 \begin{center}
$(\star + F) \odot G + (\star + F) \odot H$ 
 \end{center}
The layer and computation count numbers are 4 and 5, respectively. 
After fusion, the computation function is simplified as:
 \begin{center}
$(\star + F) \odot (G + H)$ 
 \end{center}
 Where layer and computation count numbers become  1 and 3, respectively. This process can significantly reduce the operator number and computation overhead. Compared  with prior work on loop fusion~\cite{ashari2015optimizing,bezanson2017julia,boehm2018optimizing}, the novelty of this approach is that we exploit a restricted domain of DNN execution. Thus, we  can enable more aggressive optimizations without very expensive exploration.

\begingroup
\setlength{\tabcolsep}{5.5pt} 
\begin{table*}[t!]
\centering
\setlength{\belowcaptionskip}{-10pt}
\resizebox{2\columnwidth}{!}{
\begin{tabular}{|cc|c|cccc|cccc|}
     \hline
     Framework & \multirow{2}{*}{\#FLOPs} & TFLite & \multicolumn{4}{c|}{\projectname (without layer fusion)} & \multicolumn{4}{c|}{\projectname (with layer fusion)} \\
     Device   & ~ & \makecell{CPU} & \makecell{CPU} & Speedup & \makecell{GPU} & Speedup & \makecell{CPU} & Speedup & \makecell{GPU}  & Speedup \\\hline
     DistilBERT with NAS   & 10.9G  & 188ms & 157ms & 1.2$\times$ & 237ms & 0.8$\times$ & \textbf{105ms} & \textbf{1.8$\times$} & \textbf{86ms}  & \textbf{2.2$\times$}   \\ \hline
     BERT$_{\mathrm{BASE}}$ with NAS & 21.8G & 352ms & 276ms & 1.3$\times$ & 412ms & 0.9$\times$ & \textbf{196ms} & \textbf{1.8$\times$} & \textbf{147ms} & \textbf{2.4$\times$} \\ \hline
     CANAOBERT with NAS             & 4.6G  & 98ms  & 89ms  & 1.1$\times$ & 152ms & 0.6$\times$ & \textbf{49ms}  & \textbf{2.0$\times$} & \textbf{45ms}  & \textbf{2.2$\times$} \\ \hline
\end{tabular}
}
\vskip -0.8em
\caption{Inference latency comparison of \projectname framework and TFLite on mobile CPU and GPU. All models are generated with English Wikipedia dataset. TFLite does not support BERT on mobile GPU.}
\label{tab:BERT-performance-mobile}
\end{table*}
\endgroup

\subsubsection{Polyhedral-based Code Generation}
LP-Fusion supports grouping multiple layers with varied output shapes, i.e., in the code-level, the nested loop structures of these layers may be different. Traditional compilers cannot support this kind of loop fusion well, mainly due to the complexity of such loop analysis. 
Due to space constraints, this section illustrates this complexity with an example. In this example, a trade-off exists between data locality optimization and redundant computation, thus making it difficult to select the optimal version automatically.

\begin{figure}[]
    \centering
    \setlength{\belowcaptionskip}{-0.0pt}
    \includegraphics[width=0.85 \columnwidth]{ 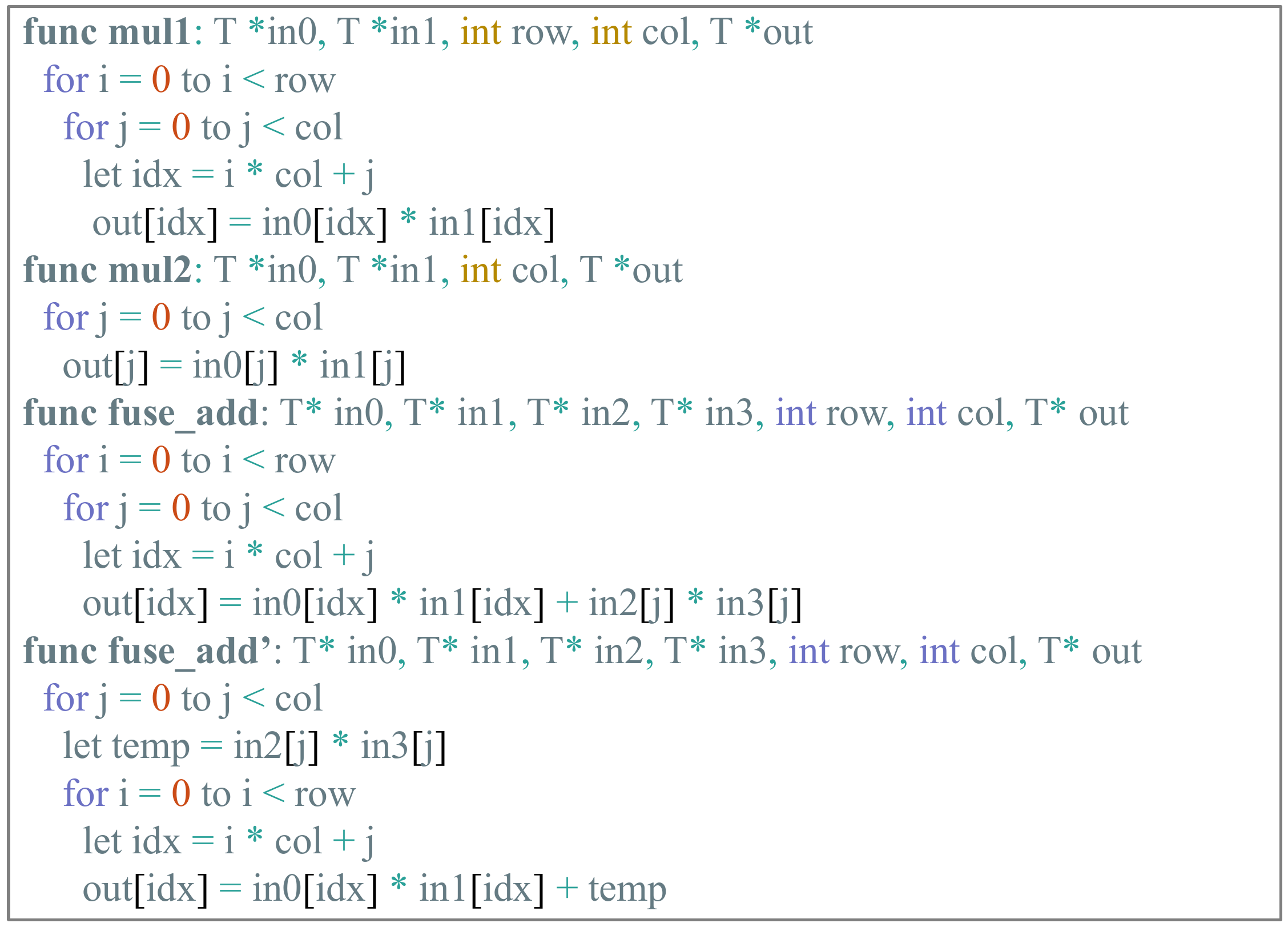}
    \vskip -0.5em
    \caption{
        An example of loop fusion.    }\label{fig:codegen_example}
\end{figure}

Fig.~\ref{fig:fusion-pattern}a and Figure~\ref{fig:codegen_example} show the example. There are three operators: {\tt Mul-1}, {\tt Mul-2}, and {\tt Add}. {\tt Mul-1} and {\tt Mul-2} take matrices $A$ and $B$ as their input, respectively. {\tt Add} takes the output of {\tt Mul-1} and {\tt Mul-2} as input to generate the result. 
The shape of matrix $A$ is M $\times$ N while the shape of B is 1 $\times$ N. 
We have two options to perform loop fusion:{\tt fuse\_add}, and {\tt fuse\_add'}. In case {\tt fuse\_add}, {\tt in2} and {\tt in3} are single dimensional arrays, so {\tt in2[j] * in3[j]} incurs redundant computation for each outer-loop iteration (except the first one). The case {\tt fuse\_add'} resolves this redundant computation with a proper loop permutation; however, it degrades the data locality because memory access for both matrices {\tt in0} and {\tt in1} becomes column-major, inconsistent with their memory storage.

To address this complexity, our compiler extends the polyhedral analysis model~\cite{WILDE1993POLYLIB} to generate both versions and employs auto-tuning to dynamically select the optimal version.
Moreover, our compiler also employs an extended polyhedral analysis model to analyze the loop structure and data dependency, transform indices, and generate optimal fused code for other loop fusion cases.

\section{Experiments and Demonstrations}

\subsection{ Training and Evaluation Setup}

Our models are trained on a server with 16$\times$ NVIDIA Tesla V100 (Volta) GPUs. We use English Wikipedia~\cite{devlin2018bert} and BooksCorpus~\cite{zhu2015aligning} to train the models and finetune on GLUE benchmark~\cite{wang2018glue}: MNLI~\cite{N18-1101}, SST-2~\cite{socher-etal-2013-recursive}, MRPC~\cite{dolan2005automatically}, STS-B~\cite{Cer_2017}, RTE~\cite{wang2018glue}, and CoLA~\cite{warstadt2018neural}. The sequence length is 128. We evaluate our framework on a Samsung Galaxy S20 cell phone with Qualcomm Snapdragon 865.
For each model, we run our framework and TFLite 100 times with 8 threads on CPU and all pipelines on GPU. 


\subsection{Demonstration on Mobile}
Figure~\ref{fig:demo} shows the interface of our real-time BERT application. Figure~\ref{fig:demo} left is the Question Answering task. Type a random question that is related to the paragraph, it will automatically highlight the answer in the test. Figure~\ref{fig:demo} right is the Text Generation task. Given a starting sentence, it can automatically generate new sentences by word. 
\subsection{Evaluation Results}

We compare the accuracy and latency of four models: BERT$_{\mathrm{BASE}}$~\cite{devlin2018bert}, MobileBERT~\cite{Sun_2020}, DistilBERT~\cite{sanh2019distilbert}, 
and CANAOBERT. 
Table ~\ref{tab:eva_distil} shows the accuracy and latency results.
We manage to significantly reduce latency compared to BERT$_{\mathrm{BASE}}$, DistilBERT, and MobileBERT on both CPU and GPU. Compared with BERT$_{\mathrm{BASE}}$, our model is 5.2$\times$ faster on CPU and 4.1$\times$ faster on GPU with 0.5-2\% accuracy loss.
Compared with MobileBERT, our model is 1.49 $\times$ faster on CPU and 1.53$\times$ faster on GPU with only 0.4-1\% accuracy decrease.

\subsection{Effectiveness of Compiler Optimizations}
We compare with a state-of-the-art framework, TFlite. 
Table~\ref{tab:BERT-performance-mobile} shows inference latency comparison results. 
TFLite only supports mobile CPU execution, and other frameworks do not support BERT models on mobile devices.
And GPU performance is unusually slower than CPU (only 0.6$\times$ speedup for CANAOBERT over TFLite on CPU). 
The fully optimized framework can achieve up to 2.0$\times$ speedup on CPU, and 2.4$\times$ on GPU, over TFLite's CPU execution. Notably, comparing to BERT$_{\mathrm{BASE}}$ on TFLite (352ms on CPU), our overall model and framework (45ms on GPU) can achieve up to $7.8\times$ speedup.

\begingroup
\setlength{\tabcolsep}{3pt} 
\begin{table}[]
\centering
\renewcommand{\arraystretch}{1.1}
\setlength{\belowcaptionskip}{-5pt}
\resizebox{1\columnwidth}{!}{
\begin{tabular}{|c|cccccc|}
    \hline
     Model &MNLI-m/mm & SST-2 & MRPC & STS-B & RTE & CoLA \\
    \hline
    BERT$_{\mathrm{BASE}}$ &84.6/83.4&93.5 & 88.9 & 85.8 & 66.4 & 52.1  \\
    \hline
    DistilBERT &81.5/81.0&92 & 85.0 & - & 65.5 & 51.3  \\
    \hline
    MobileBERT &83.3/82.6&92.8 & 88.8 & 84.4 & 66.2 & 50.5 \\
    \hline
    CANAOBERT  &82.9/82.1 & 92.6 & 88.4 & 83.5 & 65.6 & 49.2  \\
    \hline
\end{tabular}
}
\vskip -0.5em
\caption{Evaluation accuracy results on GLUE benchmark. All models are optimized with layer fusion and code generation (i.e., they already run faster than their TFLite implementation) with a fixed sequence length of 128.}
\label{tab:eva_distil}
\end{table}
\endgroup

\section{Conclusion}

We introduced a novel compression-compilation co-design framework to optimize the structures of BERT variants for mobile devices. We implemented compiler optimizations in the architecture search loop, aiming to co-optimize the model accuracy and computation resource usage. We also implemented a highly effective layer fusion method to reduce intermediate results to achieve lower latency on both mobile CPU and GPU. Further, we presented two BERT applications on mobile devices: Question Answering and Text Generation. Both can be executed in real-time with latency as low as 45ms. 

\bibliographystyle{named}
\bibliography{ijcai21}
\end{document}